\documentclass[a4paper,conference]{IEEEtran}
\ifCLASSINFOpdf
   \usepackage[pdftex]{graphicx}
\else
   \usepackage[dvips]{graphicx}
\fi
\usepackage{subfig}

\usepackage{amsmath}
\usepackage{amsfonts}
\usepackage{dsfont}

\usepackage{array}

\usepackage{bm}
\usepackage{stfloats}
\begin{document}
%

\title{Trade-offs in Top-k Classification Accuracies\\on Losses for Deep Learning}

\author{\IEEEauthorblockN{Azusa Sawada\IEEEauthorrefmark{1}, Eiji Kaneko\IEEEauthorrefmark{2}, and Kazutoshi Sagi\IEEEauthorrefmark{2}}
	\IEEEauthorblockA{\IEEEauthorrefmark{1}Biometrics Research Laboratories,\IEEEauthorrefmark{2}Data Science Research Laboratories,\\
    NEC Corporation\\
	Kawasaki, Japan\\
	Email: swd02@nec.com, e-kaneko@nec.com, ksagi@nec.com}
	}


\maketitle

\begin{abstract}
This paper presents an experimental analysis about trade-offs in top-k classification accuracies on losses for deep leaning and proposal of a novel top-k loss. 
Commonly-used cross entropy (CE) is not guaranteed to optimize top-k prediction without infinite training data and model complexities.
The objective is to clarify when CE sacrifices top-k accuracies to optimize top-1 prediction, and to design loss that improve top-k accuracy under such conditions.
Our novel loss is basically CE modified by grouping temporal top-k classes as a single class. To obtain a robust decision boundary, we introduce an adaptive transition from normal CE to our loss, and thus call it top-k transition loss. 
It is demonstrated that CE is not always the best choice to learn top-k prediction in our experiments. 
First, we explore trade-offs between top-1 and top-k (=2) accuracies on synthetic datasets, and find a failure of CE in optimizing top-k prediction when we have complex data distribution for a given model to represent optimal top-1 prediction.
Second, we compare top-k accuracies on CIFAR-100 dataset targeting top-5 prediction in deep learning. 
While CE performs the best in top-1 accuracy, in top-5 accuracy our loss performs better than CE except using one experimental setup.  Moreover, our loss has been found to provide better top-k accuracies compared to CE at k larger than 10.
As a result, a ResNet18 model trained with our loss reaches 99\% accuracy with k=25 candidates, which is a smaller candidate number than that of CE by 8.
 
\end{abstract}


\begin{IEEEkeywords}
Top-k classification, Loss function, Deep neural networks.
\end{IEEEkeywords}

%
\IEEEpeerreviewmaketitle

\makeatletter{\renewcommand*{\@makefnmark}{}
\footnotetext{All authors contributed equally.}\makeatother}
\makeatletter{\renewcommand*{\@makefnmark}{}
\footnotetext{This work has been submitted to the IEEE for possible publication. Copyright may be transferred without notice, after which this version may no longer be accessible.}\makeatother}

\section{Introduction}
Deep learning (DL) is actively being investigated and applied in various fields. Many multi-class classification studies on DL were aimed at optimizing the top-1 accuracy using cross entropy (CE) and its variants. However, in practical multi-class classification applications, we often rather exploit their top-k predictions (\textit{i.e.} $k=3,5,10$ etc.) than the top-1 prediction as candidates in combination with our recognition and other information. In such a case, top-k accuracy is an important metric that estimates whether the candidates include correct targets, which limits total performance.

Most top-k prediction studies were conducted using support vector machines (SVMs) and shallow models. Lapin \textit{et al}. \cite{lapin2016} discussed that CE is \textit{top-k calibrated}, which is the requirement to optimize top-k prediction under ideal conditions where we have enough samples to estimate true data distribution and sufficient model complexities to minimize the loss. However, it is difficult to prepare a large amount of labeled data, especially for highly complex models such as deep neural networks (DNNs). In addition, the complexity of DNNs to be trained is often limited by memory resource or inference time. Even when the DNNs have sufficient complexity, the optimization process in a manner of stochastic gradient descent (SGD), which is typically used in DL \cite{sgd}, does not usually reach the exact minima of training loss. Therefore, ideal conditions are not relevant to most cases. Our target is to investigate the superiority of losses to achieve the best possible accuracy within such a limitation of resources.

Berrada \textit{et al.} \cite{smoothTopK} suggested that non-sparse gradients of losses are important for DL and applied smoothing to a top-k SVM loss. They showed that DNNs can be trained well using their smooth loss. They also found that their loss was robust against over-fitting compared to CE when training data contain noisy labels or the amount of training data is small. Apart from the vulnerability to over-fitting, however, it remains unclear whether top-k specialized losses can be superior to CE in top-k prediction. 

This paper experimentally clarifies the conditions under which learning using CE does not optimize top-k prediction. We also propose a novel loss for the purpose of top-k classification. Our proposed loss is an extension of CE to have smaller top-k risk than the original CE. We experimentally confirmed that our loss outperforms the original CE in learning top-k predictions. 

This paper is composed as follows. Section \ref{sec2} reviews relevant literature. We explain the original CE and define our proposed losses in Section \ref{sec3}. Experiments using synthetic data and a public CIFAR-100 dataset \cite{cifar100} are discussed in Sections \ref{sec4} and \ref{sec5}, respectively. Finally, we conclude this paper in Section \ref{sec6}.


\section{Related Work}\label{sec2}
The early research on learning top-k predictions in classification by Lapin \textit{et al}. \cite{lapin2016} is an extensive analysis on top-k loss families based on SVM loss or CE. They showed CE is top-k calibrated for any $k$, which means the minimizers of CE will minimize the top-k error for all $k$ under ideal conditions. They illustrated the failure of top-1 loss in top-2 prediction using synthetic data. However, its dependency on data distributions or classifiers has not been explored in detail. In their experiments on real datasets, any of the top-k loss families has not consistently improved top-k accuracy compared to top-1 loss.

Lapin \textit{et al}. \cite{lapin2018} conducted in-depth analysis on smooth SVM loss as well as top-k loss families in an earlier study \cite{lapin2016}. They also explored single and multilabel settings in image classification. They again experimentally found that top-k losses cannot outperform softmax CE.

Recently, Yang \textit{et al.} \cite{consistency} extended top-k calibration to \textit{top-k consistency} and cover the case that top-k prediction cannot uniquely be decided. They confirmed the consistency of several existing top-k losses, and also proposed a new top-k hinge loss that satisfies top-k consistency.

Chang \textit{et al.} \cite{robustTopK} proposed a truncated re-weighted top-k SVM loss for robustness against outliers. 
Tan \cite{exactL0} suggested an algorithm for directly minimizing the top-k error as $L0$ norm.

Berrada \textit{et al}. \cite{smoothTopK} considered the difficulty specific to DL. They proposed a modification of a top-k multiclass SVM loss and a fast and stable algorithm to approximate it. They mentioned the importance of smoothness and gradient density of loss in applying it to DL for successful optimization. They proposed a loss by smoothing the max operation in a top-k multiclass SVM loss over a combination of $k$ classes and confirmed that it can be easy to minimize in DL. In their experiments, they showed that their smooth SVM top-k loss was more robust than CE to over-fitting due to noisy labels or small amount of training data. However, the superiority of their loss to CE in learning top-k prediction was not discussed.

In this paper, we explore the trade-offs between top-1 and top-k accuracies and propose a novel top-k loss starting from CE. The modification to obtain our top-k loss is consistent with conditional probability modeling by softmax outputs. This enables us to apply various techniques used with CE or probabilistic treatments.

\section{Method}\label{sec3}
In this section, we first explain our problem and notations. We then introduce CE and our proposed loss.

\subsection{Problem statements}\label{sec31}
We consider multi-class classification of $N$ classes with data space $\mathcal{X}$ and label space $\mathcal{Y} = \{1, \cdots, N\}$. Classifiers receive input $x \in \mathcal{X}$ and they return $N$-dimensional vector $\bf{s} \in \mathds{R}^N$ or normalized values $\mathbf{q}$ after softmax operation as their outputs. Top-k prediction for $(k=1, \cdots, N-1)$ is a class-label set corresponding to top-k largest components of outputs. We refer to the index sorted in descending order of $\mathbf{q}=(q_1,q_2,..,q_N)$ as $[i]$ for $i=1, \cdots, N$, then top-k prediction can be written as
\begin{align}
\mathcal{C}_k(\bm{q})= \{ y_{[i]} \in \mathcal{Y} | 1 \leq i \leq k \}.
\label{eq-topkpred}
\end{align}
This can cause ambiguity of sorting when an output contains equal components.

Top-k error is now written as an indicator if true label $y$ is not a part of the top-k prediction $\mathcal{C}_k(\mathbf{q})$
\begin{align}
err_k(\bm{q},y) = \mathds{1}[y \notin \mathcal{C}_k(\bm{q})],
\label{eq-topkerr}
\end{align}

where $\mathds{1}$ denotes the indicator function over Boolean statements; 1 if true, 0 if false. Top-k risk is the expectation value of this error. To minimize the risk, we often alter it by a surrogate loss, such as CE or hinge loss, in the optimization.

\subsection{Cross entropy loss}\label{sec32}

CE is the typical surrogate loss of top-1 risk and can be written using the one-hot representation $\textbf{t}$ of $y$ as 
\begin{align}
\mathcal{L}_{CE} = -\log q_y = -\sum_{i=1}^{N}{t_i \log q_i}.
\label{eq-ce1}
\end{align}

The gradients of this function with softmax are dense so that gradient descent in DL works well.

CE can also be used to minimize the distance between probability distributions. The normalized outputs $q_{i}$s for $i$ are non-negative and sum to 1 so that $q_y$ can model the conditional probability of $y$ given $x$. In this sense, CE can be seen as a $\textbf{q}$-dependent part of KL-divergence $D_{KL}$ between $\textbf{q}$ and target distribution $\textbf{t}$,
\begin{align}
\mathcal{L}_{CE}(\bm{q},\bm{t}) = -\sum_{i=1}^{N}{t_i \log q_i} = D_{KL}(\bm{t}||\bm{q})+H(\bm{t}),
\label{eq-ce2}
\end{align}
where $H(\textbf{t})$ is the entropy of $\textbf{t}$. Thus, CE shows how outputs $\textbf{q}$ differ from $\textbf{t}$ when these values are considered as probability distributions.

CE can be used as the surrogate of top-k risk because it is top-k calibrated for $k>1$ as well as $k=1$ \cite{lapin2016}. This is a preferable property for surrogate loss, which indicates the success in top-k performance of models trained using this loss in DL with large datasets.

However, the calibration does not directly matter for the solutions except minimizers of the loss under ideal resource in data and the complexity of classifiers. Since the optimization process does not always reach such minimizers even in the end of training, it is important to prefer solutions with as small a risk as possible.

\begin{figure}[!t]
    \centering
    \includegraphics[width=2.5in]{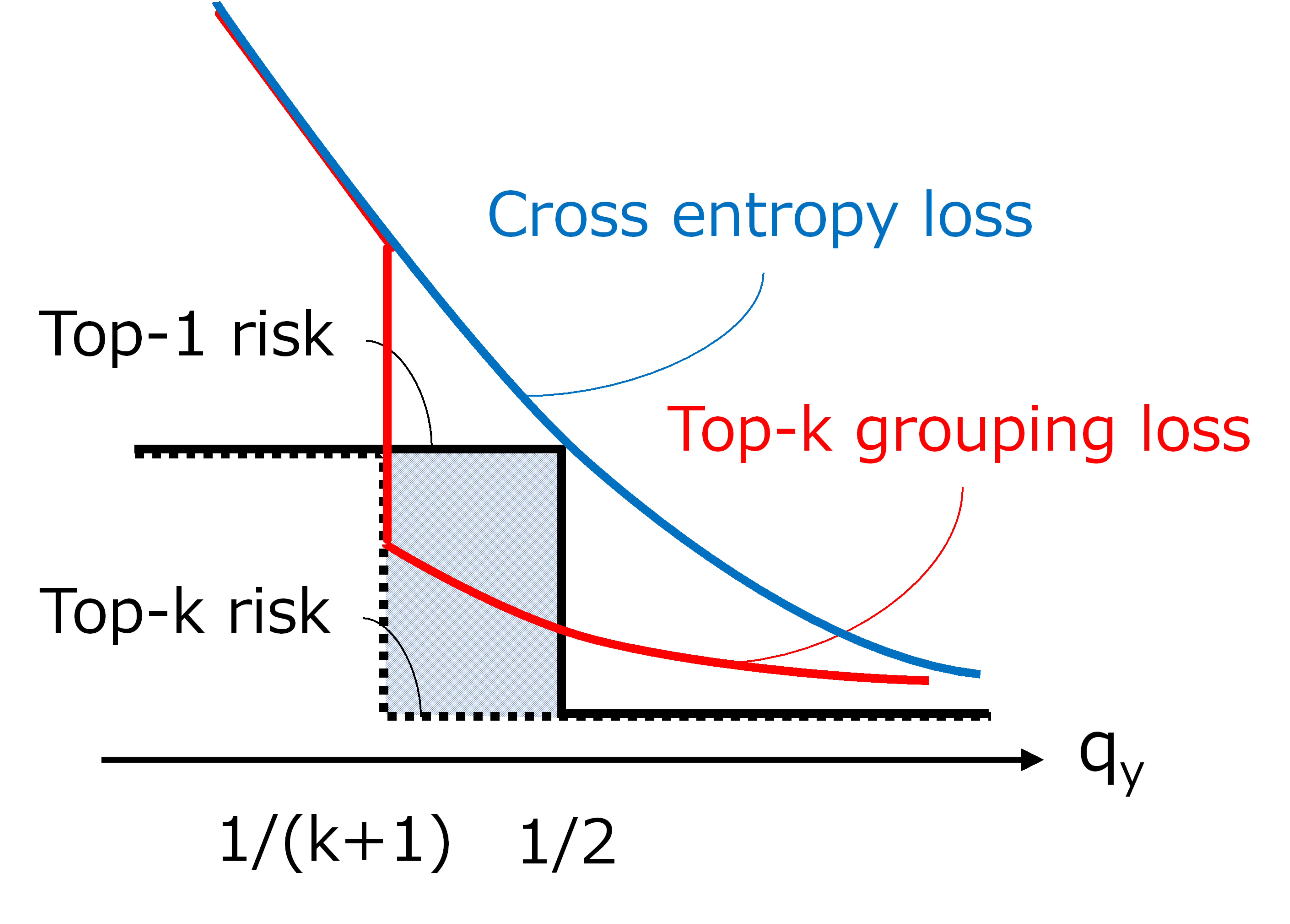}
    \caption{Schematic view of losses and risks for single data point. Horizontal axis shows output $\textbf{q}$ for true label or most probable class $y$. CE and the proposed top-k grouping loss are plotted in blue and red curves, respectively. Black steps in solid and dotted lines show top-1 risk and top-k risk, respectively. Note that this is1-dimensional slice in $N$-dimensional space of $\textbf{q}$.}
\label{fig:concept_loss}
\end{figure}

There is an extra gap between CE and top-k risk since top-k risk can be zero depending on top-2 to top-k predictions even if top-1 prediction is incorrect. Figure \ref{fig:concept_loss} shows the relation between CE, top-1 risk and top-k risk for a certain data point. CE penalizes top-k correct predictions similarly to incorrect predictions when it is incorrect in top-1 predictions. This leads to a sacrifice in top-k correct prediction on samples difficult to predict correctly in top-1 prediction, instead of achieving top-1 correct prediction on other samples, to minimize the average of losses over all training samples. Therefore, CE is potentially not the best choice to obtain the best top-k accuracy of all possible predictions within practical limitations.

\subsection{Top-k transition loss}\label{sec33}
\subsubsection*{\textbf{Top-k class grouping}}\label{sec331}

To obtain a smaller gap onto top-k risk with $k > 1$, we extend CE to derive our top-k loss. Since top-k risk is independent on a ranking in predicted top-k classes, we extend CE to a form independent of the rank. In the extended loss, we do not use original outputs $q_{[i]}$ and target $t_{[i]}$ for the predicted top-k classes $i=1,...,k$ individually. We group these classes as a single class, referred to as top-k union, and use only summations of $q_{[i]}$ for classes in the group instead.

The extended loss, which we call top-k grouping loss $\mathcal{L}_k$, is defined as CE loss calculated for the top-k union and the remaining classes as below,
\begin{align}
\mathcal{L}_{k} = -\left( \sum_{i=1}^{k}{t_{[i]}} \right) \log \left(\sum_{i=1}^{k}{q_{[i]}}\right) - \sum_{i=k+1}^{N}{t_{[i]}\log q_{[i]}}.
\label{eq-topK}
\end{align}

This form ignores the class difference in top-k prediction $\mathcal{C}_k(\textbf{q})$ and permits the confusion in top-k classes. From the viewpoint of probability mentioned in the previous section, the first term is for a probability that an input belongs to one of predicted top-k classes.

This loss has discontinuity at the point where $y$ is getting into the top-k prediction $\mathcal{C}_k(\bf{q})$ because the remaining term changed the input of the logarithm from $q_y$ to the summation of $q_y$ and other $q_{[i]}$s in $\mathcal{C}_k(\bf{q})$. In gradient descent optimization, we can define the gradient on the discontinuous point, for example, as the gradient from the second term.

\subsubsection*{\textbf{Loss transition from CE to top-k grouping loss}}\label{sec332}
The $\mathcal{L}_k$ do not control which classes are contained in top-k predictions, which has an impact on the final test performance. In the early stage of training, the model prediction is uncertain and top-k grouping can be decided at random. There are also possible ways of top-k grouping that can minimize $\mathcal{L}_k$, especially for data distribution in which the classes are well-separated from each other. However, $\mathcal{L}_k$ do not take into account the order of the candidates. Therefore, training only by $\mathcal{L}_k$ will not lead to moderate top-k prediction even if top-k error is minimized for training data. It will be difficult to obtain robust decision boundaries by randomness from possibility, which can be seen as unstable performance in our experiments. 

To avoid such risk, we define a loss that uses CE loss at the beginning of training and gradually makes a transition from CE loss to $\mathcal{L}_k$ as follows, 
\begin{align}
\mathcal{L}_{CE,k} = \left(1-\sum_{i=1}^{k}{q_{[i]}}\right)\cdot \mathcal{L}_{CE} + \left(\sum_{i =1}^{k}{q_{[i]}}\right) \cdot \mathcal{ L}_{k}.
\label{eq-transition}
\end{align}

Since $\sum_{i=1}^{k}{q_{[i]}}$ is small for uncertain predictions and becomes closer to 1 as training continues, the transition from CE to our $\mathcal{L}_k$ can be done adaptively with this form with Eq. \ref{eq-transition}. We call this ``top-k transition loss" later in this paper. 

\section{Experiments on Synthetic Data}\label{sec4}

In this section, we explore the possible failure of CE in optimizing top-k prediction by comparing the performance of CE and our top-k losses on simplified settings. Note that we call the top-k grouping loss and the top-k transition loss as top-k losses in this section. To clarify the conditions under where the failure of CE occurs, we investigate the dependency on data distribution and model complexity.

First, we focused on the data distribution dependency with a fixed model. In Experiment 1, we used the simple data distribution, where each class comes from a single normal distribution, located on a circle, with different inter-class overlap. We extended this to a more complex distribution in Experiment 2, by adding a minor mode to each class distribution at a different distance from the major mode. 

Next we examined the dependency on model complexity in Experiment 3 using the data distribution that was found to cause sub-optimal top-k performance of CE in the preceding experiments.
Detailed setups and discussions for each experiment are described in the following subsections.

\subsection{Experiment 1: On class confusion}\label{sec41}
In this experiment, we examined the performance of CE and the top-k losses on different strengths of inter-class confusion that increase the top-1 risk of the best possible prediction, which can be trained with a model having infinite complexity. We set $k=2$ in six classes.

\subsubsection*{\textbf{Settings}}

We used the synthetic data distribution that consists of normal distributions with standard deviation $\sigma$ on the circle of radius 1, each mode belongs to each class respectively. Let $\theta$ be the parameter that indicates the position on the circle as the angle from one of the positive axes in data space. We have six classes and their centers are placed at equal spaces 60$\,^{\circ}$ in $\theta$. Parameter $\sigma$, which controls confusion strength, is changed to 10$\,^{\circ}$, 15$\,^{\circ}$, 20$\,^{\circ}$, 25$\,^{\circ}$, and 30$\,^{\circ}$. The number of confusing classes at each data point varies as $\sigma$ changes. We sampled 300 data points per class from the distribution both for training and evaluation data. 

\begin{figure*}[!t]
    \centering
    \subfloat[Sampled dataset]{\includegraphics[width=2.3in]{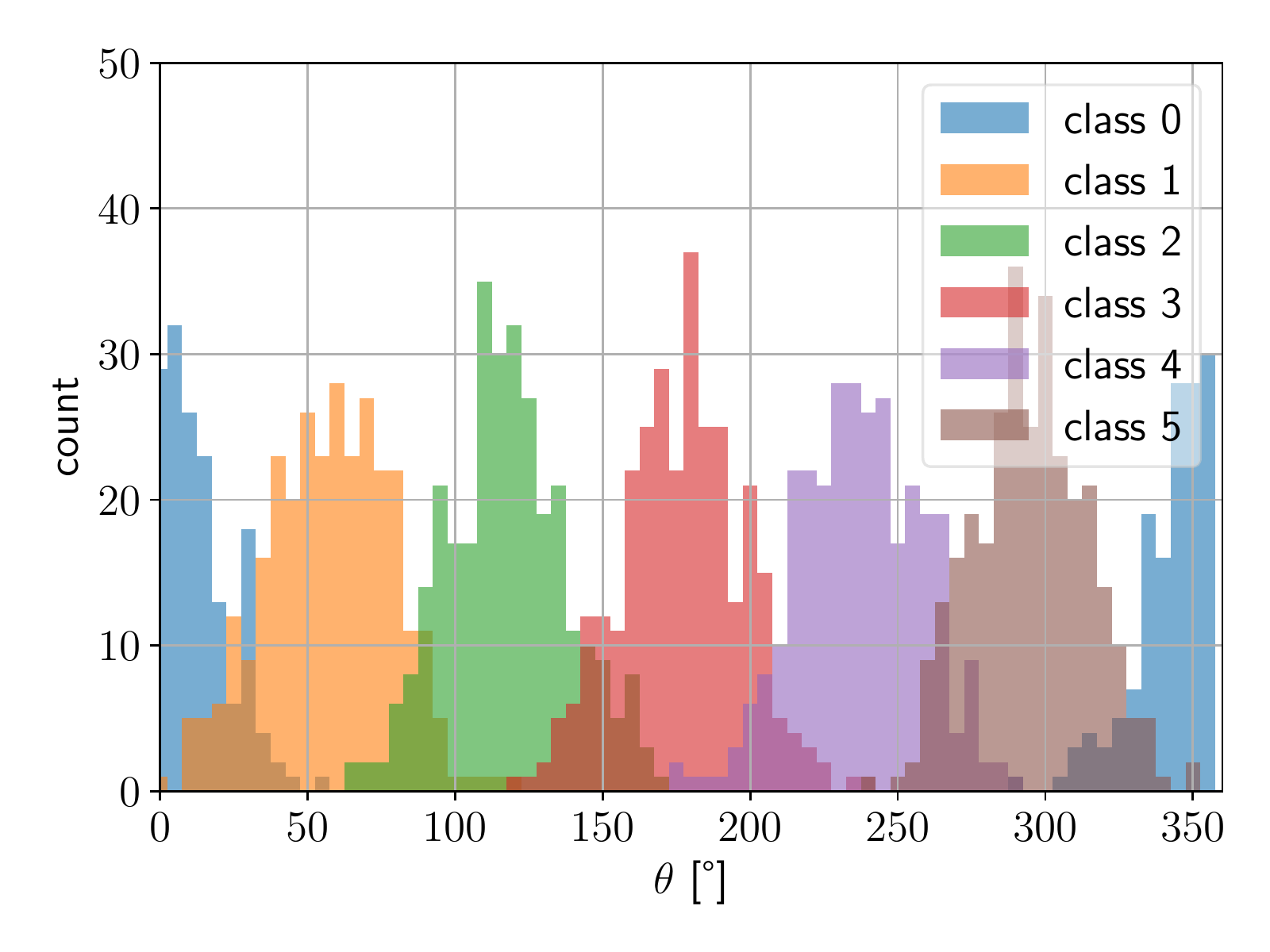}
    \label{fig:sim_dataset_a}}
    \hfil
    \subfloat[Top-1 accuracy]{\includegraphics[width=2.3in]{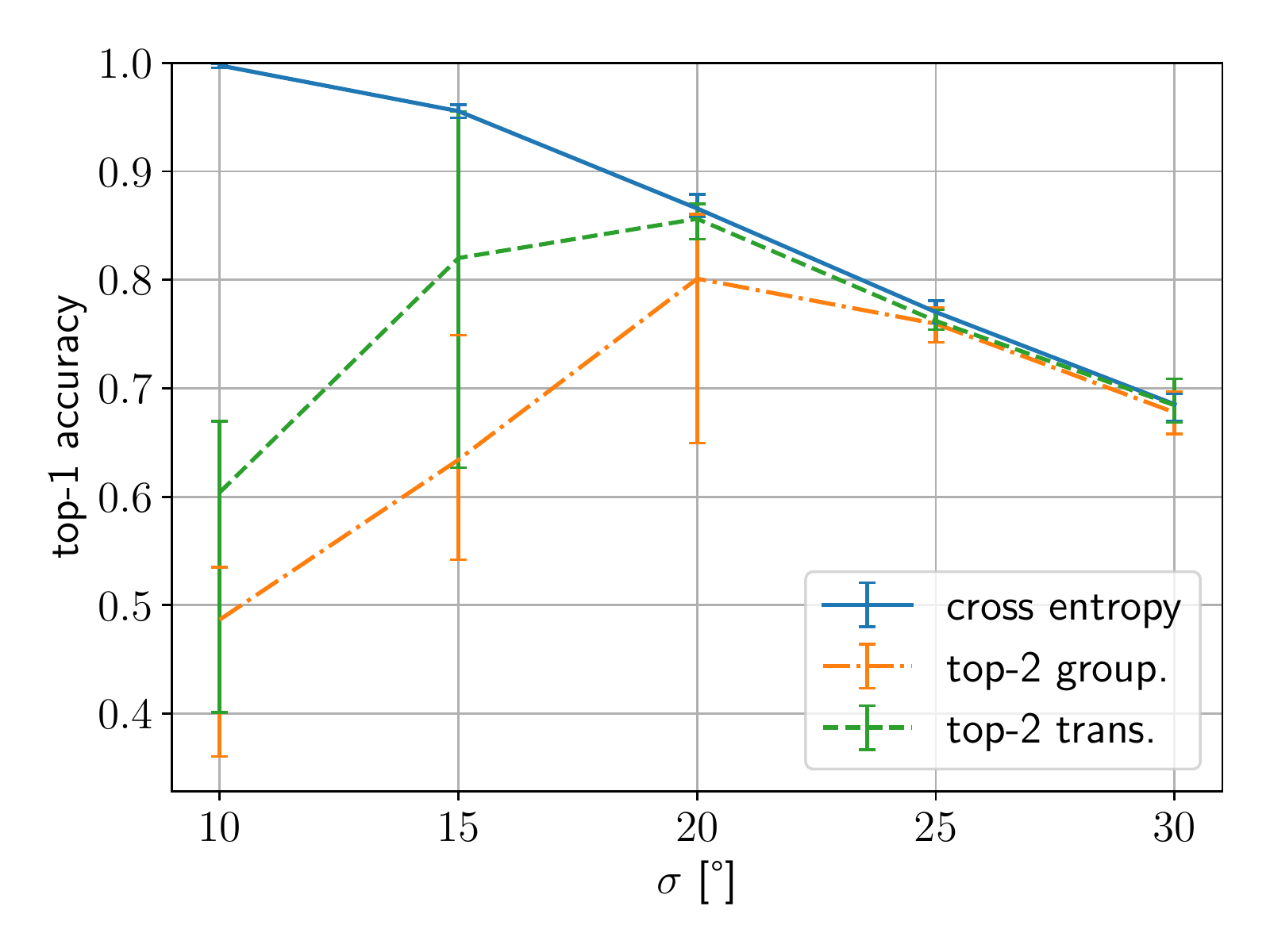}
    \label{fig:sim1a}}
    \hfil
    \subfloat[Top-2 accuracy]{\includegraphics[width=2.3in]{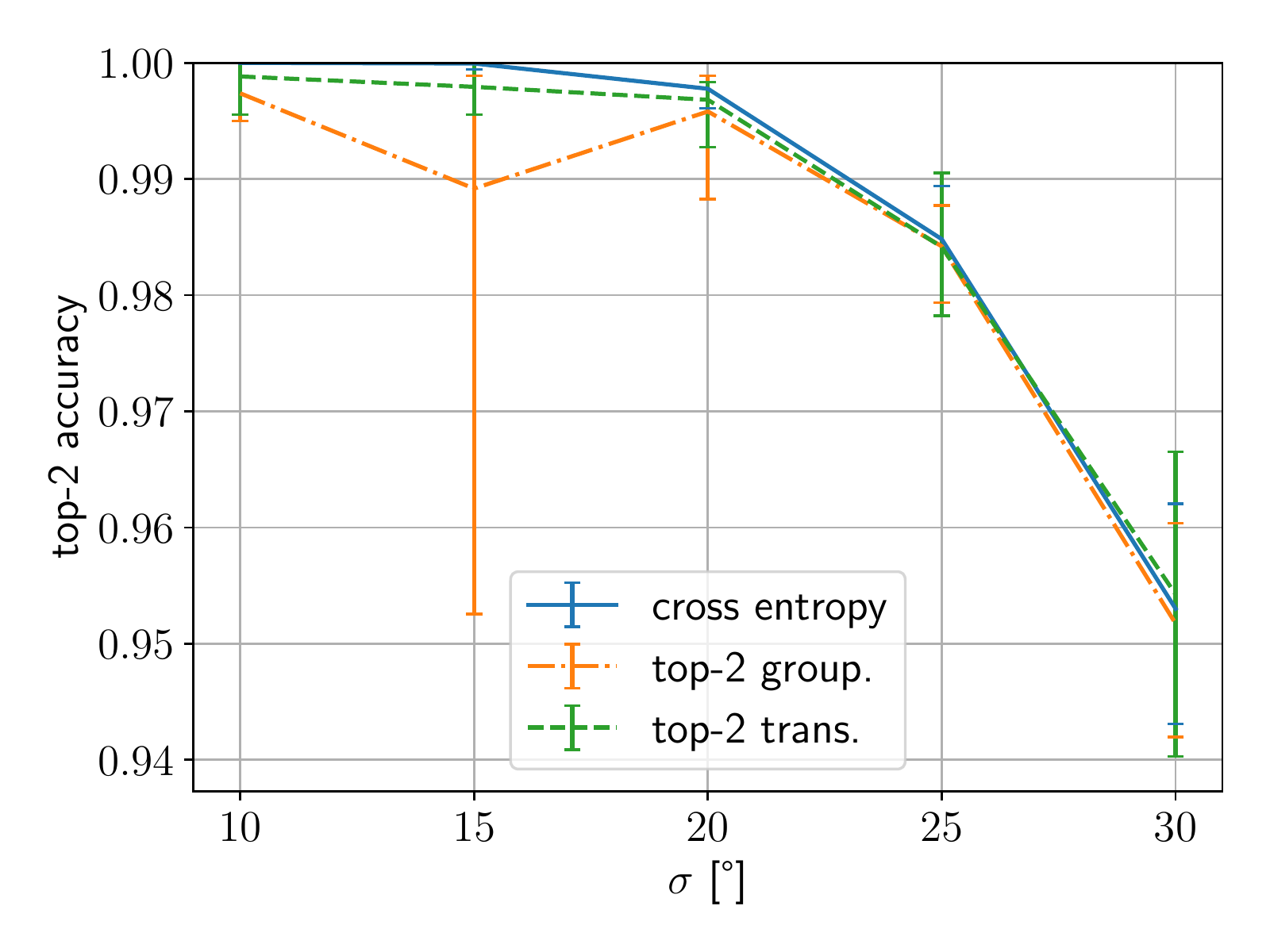}
    \label{fig:sim1b}}
    \caption{Results from Experiment 1 for different widths of single class mode $\sigma$. (a) Example of sampled data $\sigma=20\,^{\circ}$. (b) Top-1 and (c) top-2 accuracies by each loss.}
\label{fig:sim1}
\end{figure*}

Figure \ref{fig:sim_dataset_a} shows an example of the data distribution when $\sigma$ was set to 20$\,^{\circ}$. In this case, there are about two confusing classes at any data point since the tail at 99.7 \% of each normal distribution reaches the center of the neighbor class distribution. Note that inter-class confusion decreases the best possible top-1 accuracy for each data point up to the ratio of class.

For a classifier, a neural network composed of a two-dimensional input layer, a fully connected layer to six dimensions without bias terms, and an activation layer with softmax function was used. The network was trained for 1,000 epochs.

We report on the test accuracies averaged over ten runs with different random seeds in sampling the datasets and training the networks. The top-1 and top-2 performances for CE, the top-2 grouping loss, and the top-2 transition loss were compared at each $\sigma$. 


\subsubsection*{\textbf{Results}}

Figure \ref{fig:sim1} shows the top-1 and top-2 accuracies of the network trained by each loss for different $\sigma$. The error-bar shows maximum and minimum accuracies in the trials. In top-2 accuracy as well as in top-1 accuracy, CE obtained almost the best  of all losses in this settings. We can see in top-1 and top-2 accuracies a tendency that CE performed better than our losses, especially for $\sigma<20\,^{\circ}$, where inter-class overlaps are small. For such a non-overlapping region, the top-2 losses cannot uniquely determine top-2 class combination to predict. This leads to not only low top-1 accuracy but also no robust boundary of the top-2 combination in contrast to CE. For this reason, the top-2 accuracy of our losses degrades on average when there is a discrepancy between training and test data. This can be seen as large error-bar on $\sigma=15\,^{\circ}$ in Fig. \ref{fig:sim1b} due to large degradation in a few trials.

The performance of the top-2 losses asymptotically approached CE for $\sigma\geq 20\,^{\circ}$, where more than two classes significantly overlap. With large overlaps, the top-2 losses can find which combination of top-2 classes to predict over almost all data points and decision boundaries similar to that of CE. As a result, accuracies of the network trained the top-2 losses approached CE for large $\sigma$.


\subsection{Experiment 2: On multi-modality of distribution}\label{sec42}
The failure of CE in top-2 performance was not found in Experiment 1 where confusion strength was controlled with the optimal prediction boundary fixed. In this experiment, we considered a more complex data distribution by adding a minor mode to each class distribution. We set $k=2$ in six classes, as in Experiment 1.

\subsubsection*{\textbf{Settings}}
We used the same data distribution as in Experiment 1 with $\sigma=10\,^{\circ}$ as major modes. Unlike experiment 1, each class had an additional minor mode of normal distribution2 times less frequent as the major mode with the same width. Each minor mode is located with shift in $d$ from the major mode of the same class, and $d$ is changed to 30$\,^{\circ}$, 45$\,^{\circ}$, 60$\,^{\circ}$, 75$\,^{\circ}$ and 90$\,^{\circ}$. Dataset sampling, training, and evaluation were carried out in the same manner as in Experiment 1. The same model architecture as also used. The top-1 and top-2 performances for CE, the top-2 grouping loss, and the top-2 transition loss were compared at each $\sigma$. 

\begin{figure*}[!t]
    \centering
    \subfloat[Sampled dataset]{\includegraphics[width=2.3in]{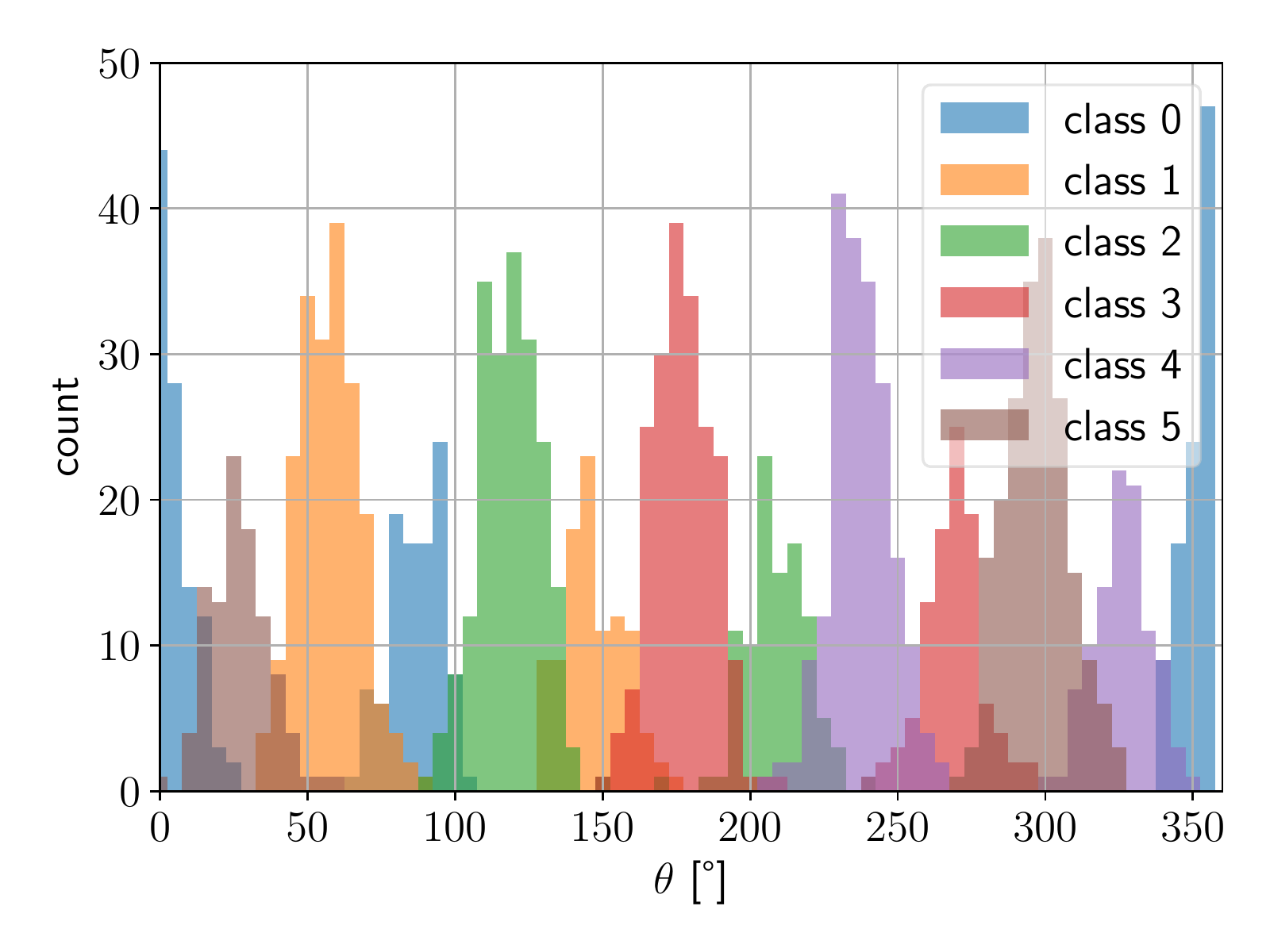}
    \label{fig:sim_dataset_b}}
    \hfil
    \subfloat[Top-1 accuracy]{\includegraphics[width=2.3in]{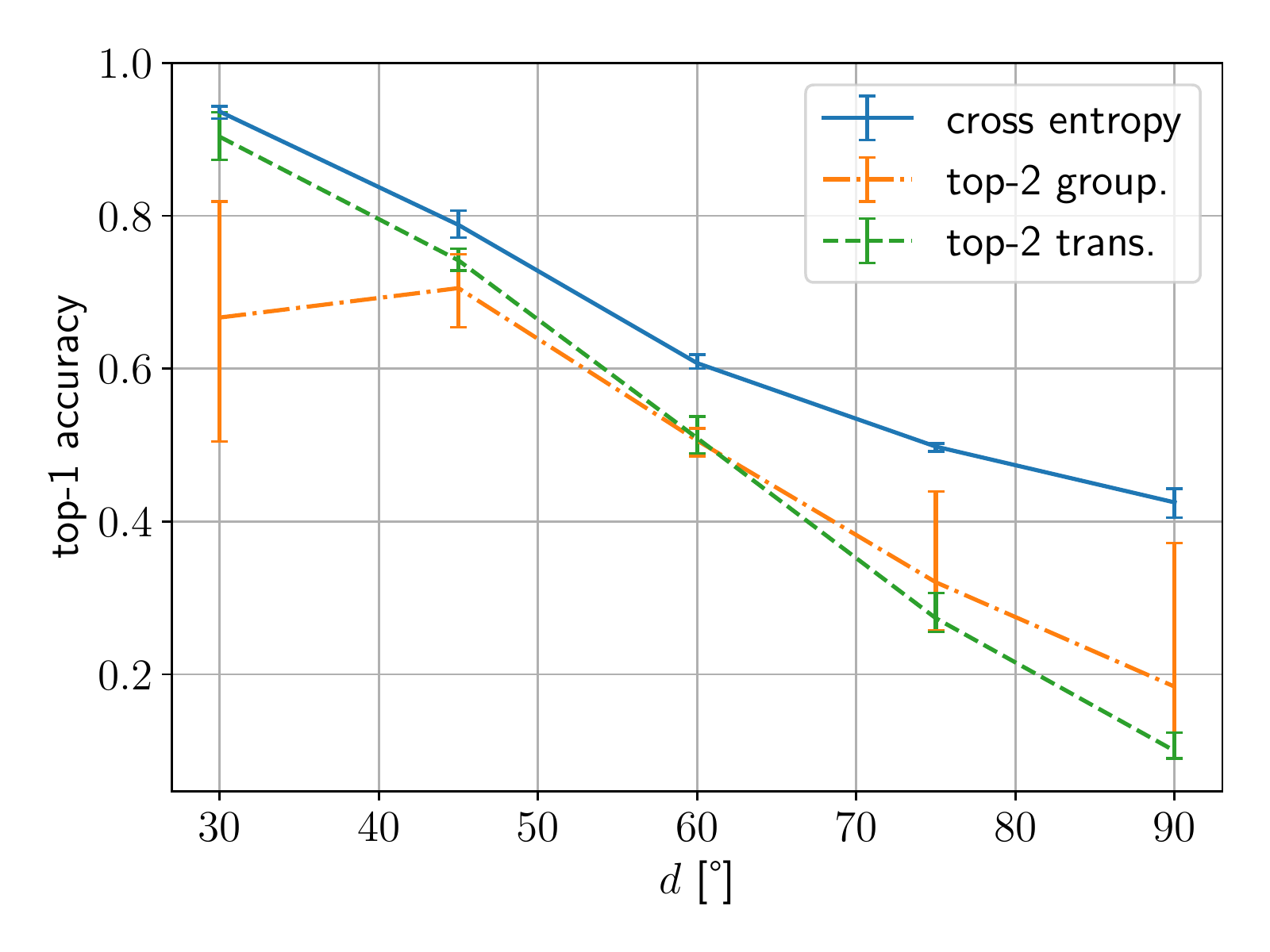}
    \label{fig:sim2a}}
    \hfil
    \subfloat[Top-2 accuracy]{\includegraphics[width=2.3in]{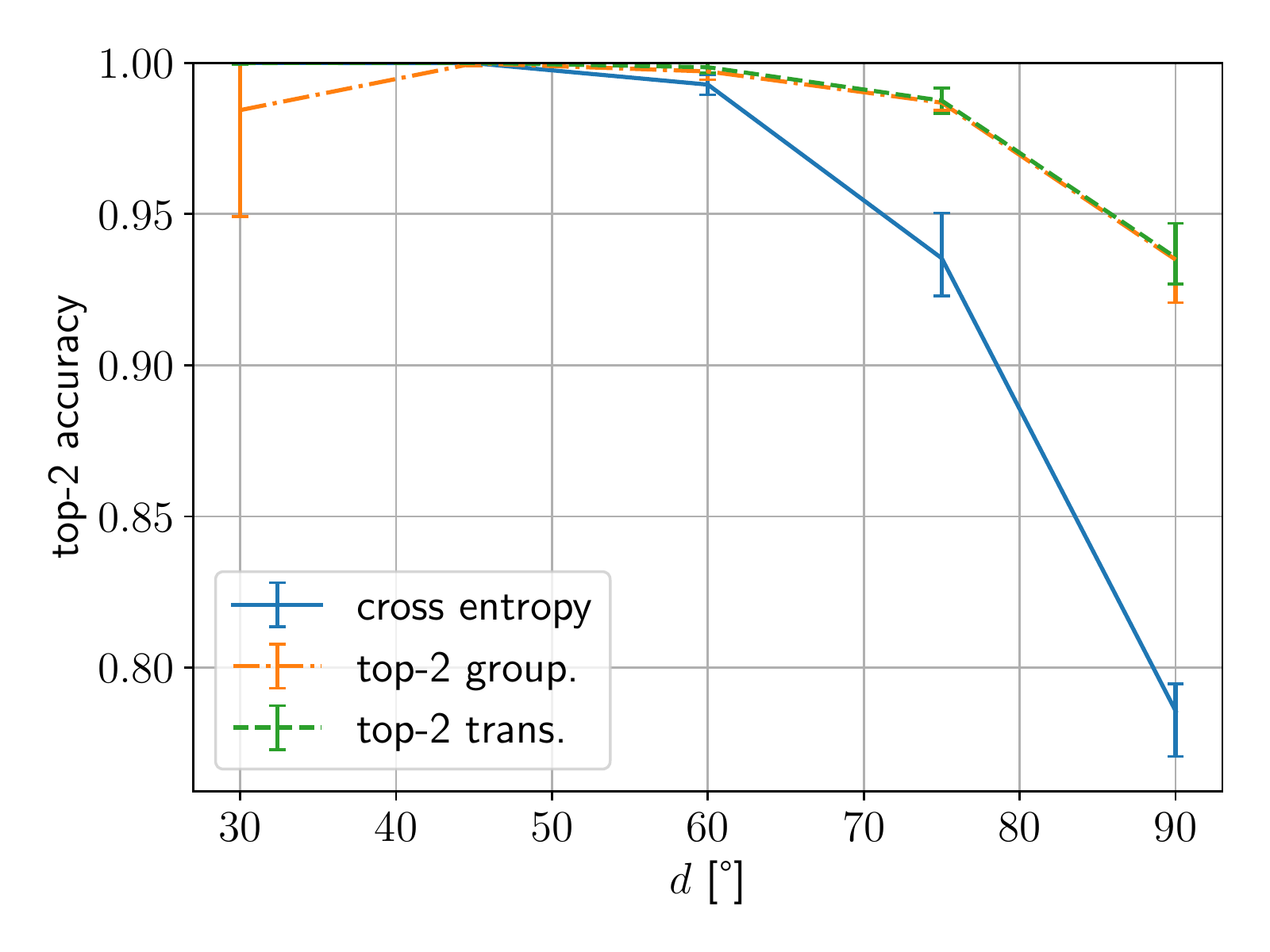}
    \label{fig:sim2b}}
    \caption{Results from Experiment 2 for different inter-mode distance $d$. (a) Example of sampled data $d=90\,^{\circ}$. (b)Top-1 and (c) top-2 accuracies by each loss.}
\label{fig:sim2}
\end{figure*}

Figure \ref{fig:sim_dataset_b} shows an example of the data distribution for this experiment for $d=90\,^{\circ}$. Under this condition, each minor mode was located between the major modes of the nearest and next nearest neighbor classes. The minor modes make additional split regions of optimal prediction.



\subsubsection*{\textbf{Results}}

Figure \ref{fig:sim2} shows the top-1 and top-2 accuracies of the network trained by each loss for different $d$. We can see the trade-off between top-1 and top-2 accuracies for $d>60\,^{\circ}$.
From region $d<45\,^{\circ}$, CE is superior to the top-2 losses both in top-1 and top-2 accuracies. The distribution of each class only becomes asymmetric due to the minor mode in this region, which can be understood in the same manner as in Experiment 1 with small $\sigma$.
On the other hand, we can see that CE is inferior to the top-2 losses in top-2 accuracy for $d\geq 60\,^{\circ}$, which is the case in which CE failed to obtain the best possible top-2 accuracy.



\subsection{Experiment 3: On model complexity}\label{sec43}
In this experiment, we explored the dependency on model complexities that were fixed in the two experiments above. 

\subsubsection*{\textbf{Settings}}
We used the same data distribution as in Experiment 2 for $d=90\,^{\circ}$, where CE failed to fit the optimal top-2 prediction. Dataset sampling, training, and evaluation were carried out in the same manner as in Experiments 1 and 2.

The model architecture was different from that used in the above experiments, a hidden layer followed by ReLU activation were inserted after the input layer. To control model complexity, we changed the number of nodes $m$ for the hidden layer, i.e., 2, 4, 8, 16, and 32. The average top-1 and top-2 accuracies for CE, the top-2 grouping loss, and the top-2 transition loss were compared for each $m$.


\begin{figure*}[!t]
    \centering
    \subfloat[Top-1 accuracy]{\includegraphics[width=2.5in]{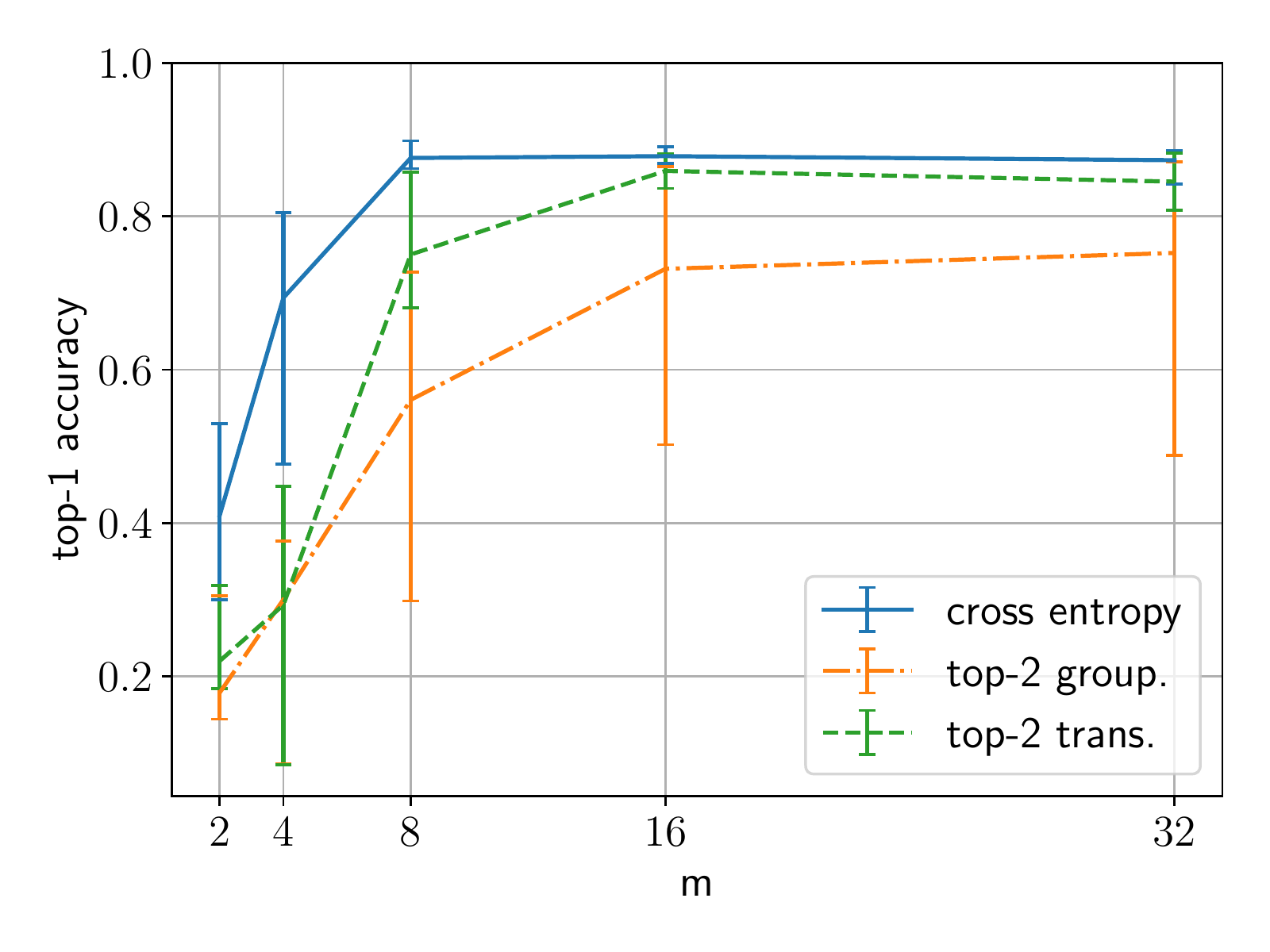}%
    \label{fig:sim3a}}
    \hfil
    \subfloat[Top-2 accuracy]{\includegraphics[width=2.5in]{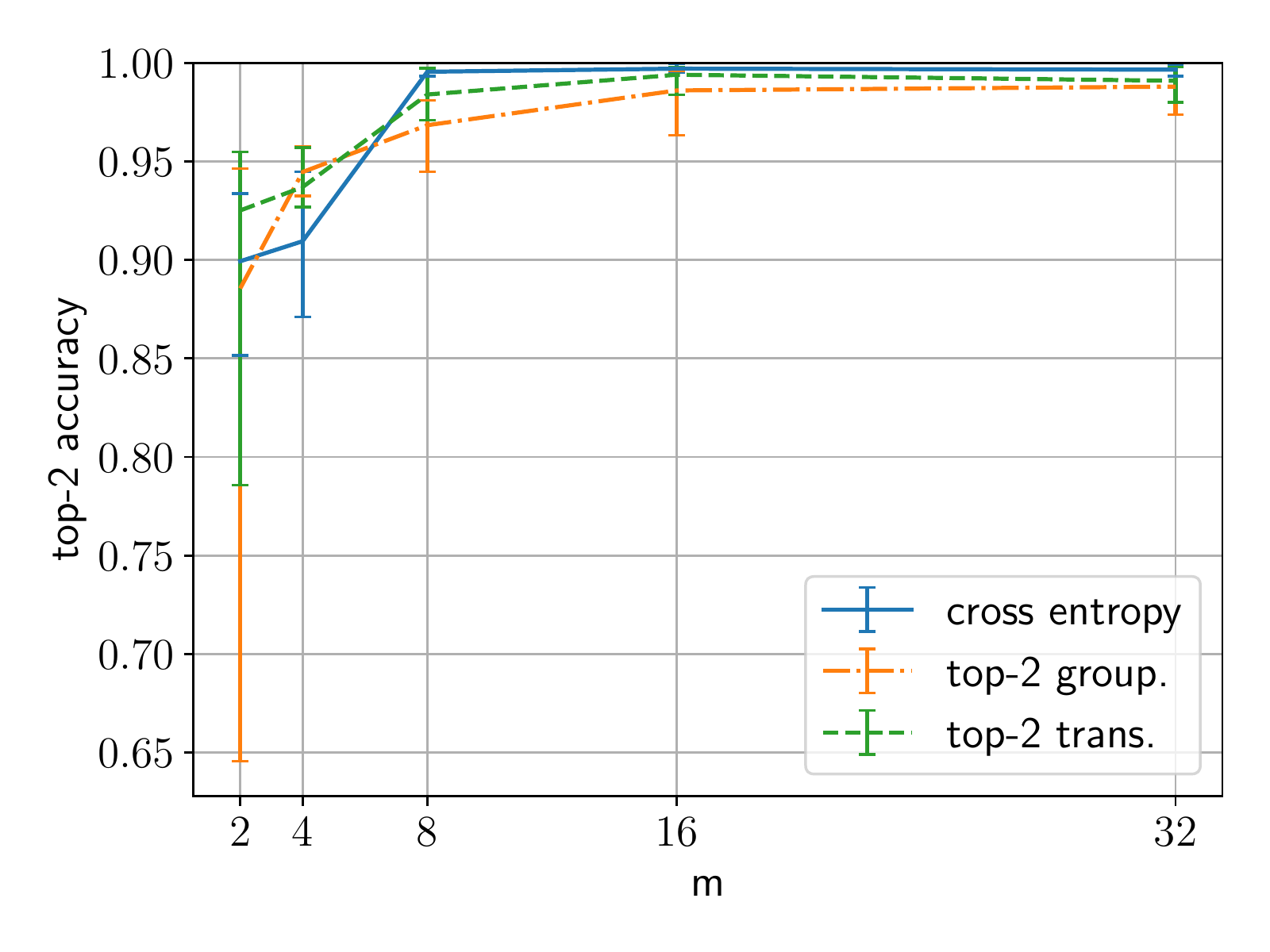}%
    \label{fig:sim3b}}
    \caption{Results from Experiment 3 for different number of nodes in hidden layer $m$. (a)Top-1 and (b) top-2 accuracies for each loss.}
\label{fig:sim3}
\end{figure*}

\subsubsection*{\textbf{Results}}

Figure \ref{fig:sim3} shows the top-1 and top-2 accuracies of the network trained by each loss for different $m$. The top-2 performance by CE was not the best of all losses for low complexity $m\leq4$, the same as in Experiment 2. For higher complexity $m\geq 8$, CE achieved the best performance both in top-1 and top-2 accuracies.


\subsection{Visualization of resulting predictions}\label{sec44b}

\begin{figure*}[!t]
    \centering
    \subfloat[Experiment 2 for $d=90\,^{\circ}$ ]{\includegraphics[width=3.0in]{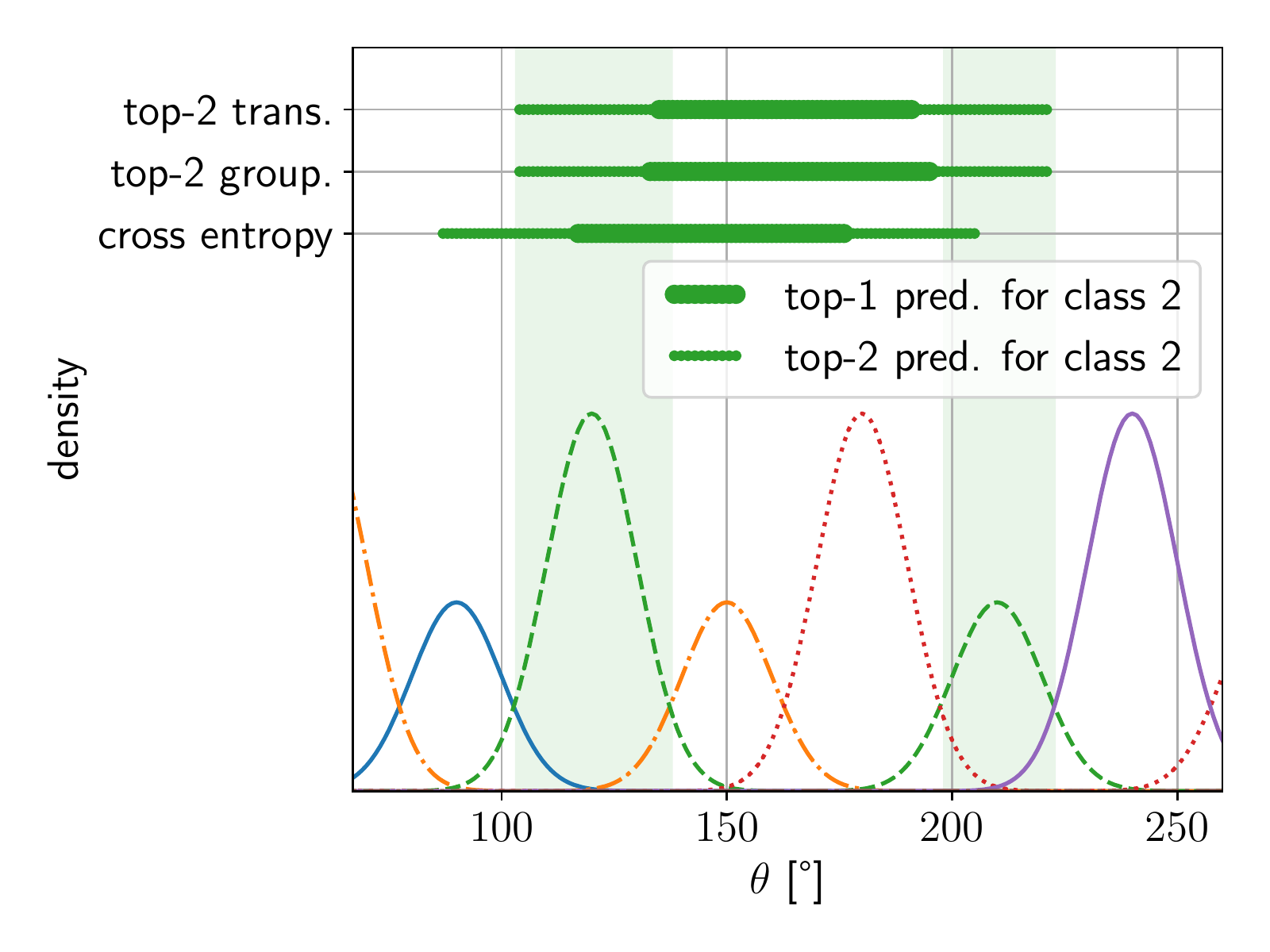}
    \label{fig:sim_analysis_a}}
    \hfil
    \subfloat[Experiment 3 for $m=8$]{\includegraphics[width=3.0in]{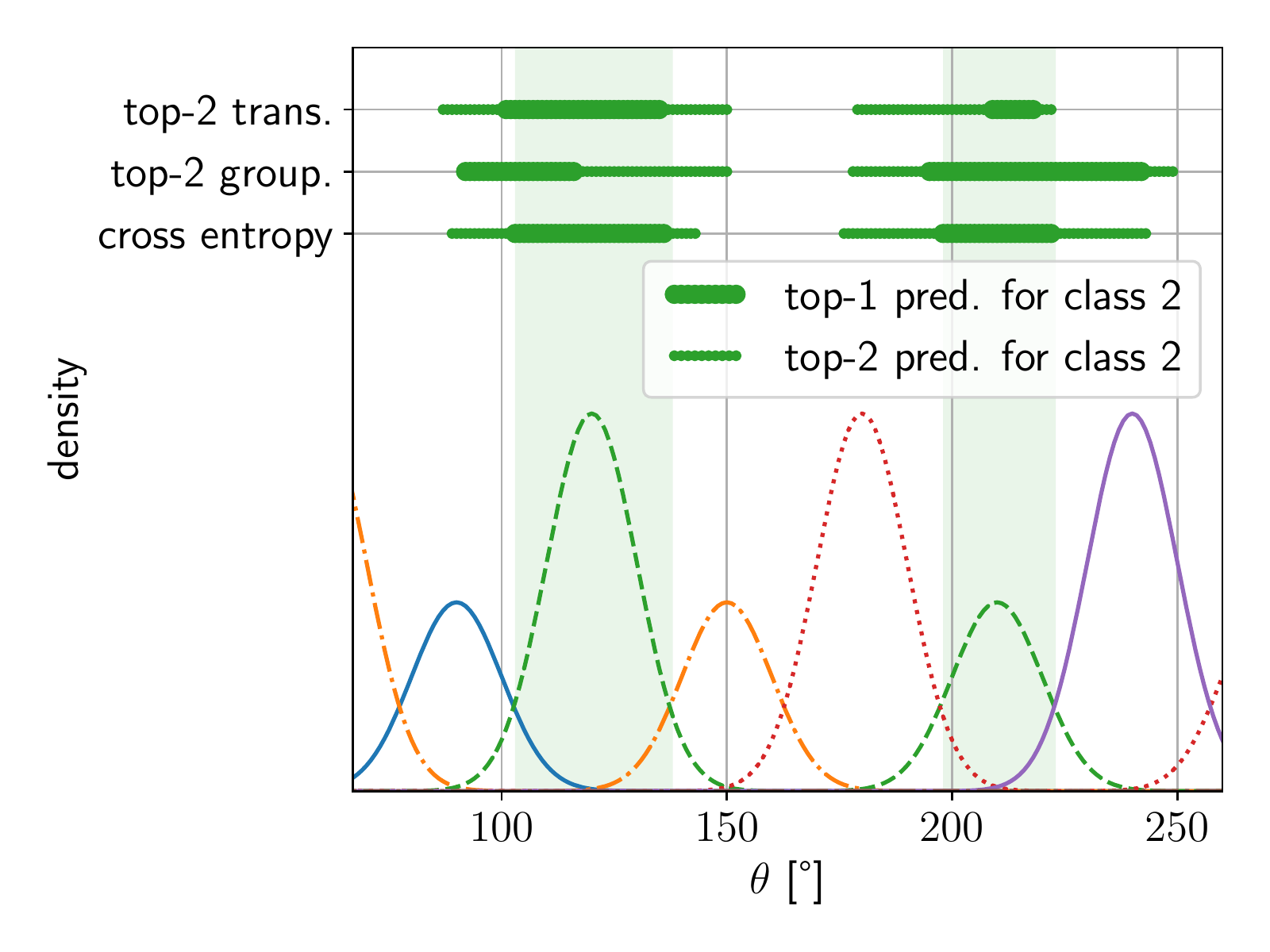}
    \label{fig:sim_analysis_b}}
    \caption{Visualization of data distribution and resulting predictions by each loss (a) in Experiment 2 for $d=90\,^{\circ}$ and (b) in Experiment 3 for $m=8$. Predictions for class 2 are shown with bars at top of each figure.}
\label{fig:sim_analysis}
\end{figure*}

To take a closer look at the failure of CE in top-2 accuracy, Figures \ref{fig:sim_analysis_a} and \ref{fig:sim_analysis_b} visualize expected data distributions and the resulting predictions for a case of $d = 90 \,^{\circ}$ in Experiments 2 and 3, respectively. Lines at the bottom of each figure are the data distributions. Focusing on the class indicated as the green line, the shaded area shows where the class is the most frequent. Thick and thin horizontal bars above the distributions show top-1 and top-2 predictions of the target class for each loss, respectively.

Figure \ref{fig:sim_analysis_a} is for a network without any mid-layers. The top-2 grouping loss and the top-2 transition loss cover all the green area, while CE does not. In contrast, CE has a better top-1 prediction than our loss in a major mode. These results indicate that CE does not optimize top-2 prediction when data has a more complex distribution than the model complexity. Moreover, our losses can perform better in top-2 prediction sacrificing top-1 prediction.

Figure \ref{fig:sim_analysis_b} is same as Fig. \ref{fig:sim_analysis_a} except for the case containing a mid-layer with nodes of $m=8$ in Experiment 3. In contrast to Fig. \ref{fig:sim_analysis_a}, there are two split areas of top-1 prediction for all losses corresponding to a true data distribution. All losses cover the most frequent region for the target class, that is, they obtained almost optimal top-2 predictions. Moreover, top-1 prediction of CE represents optimal top-1 prediction well. This shows that the network with nodes of $m \ge 8$ has sufficient complexity to fit optimal top-1 prediction for this data distribution, unlike the network in Experiment 2. This suggests that CE can obtain the best possible top-2 predictions when we use models complex enough to fit optimal top-1 prediction.

\subsection{Brief summary}\label{sec44}
We summarize our findings from the experiments on synthetic data.

According to the results from Experiments 1 and 2, CE can suffer from the trade-off between top-1 and top-k performances when the data distribution has complex structures. The inter-class confusion is not an essential component that caused top-k failure of CE in Experiment 1 while the optimal risk increased. We found that CE does not perform well in top-k accuracy when the distribution of a certain class has multiple modes that leads to a complex decision boundary to achieve optimal top-1 prediction, which can be observed in practical data.

Experiment 3 showed that such trade-off does not occur when using more complex models that can fit the optimal top-1 decision boundary for a given data distribution. It is suggested that CE leads to almost optimal top-k prediction as well as top-1 prediction with enough model complexity. 


After all, CE loss is not always an optimal choice to learn top-k prediction and it happens for example when a model does not have sufficient complexity to fit top-1 prediction of a data distribution. Our losses can achieve better top-k accuracy under such conditions. 








\section{Experiment on CIFAR-100}\label{sec5}
We investigated the performance for each loss on more realistic data with deep convolutional neural networks.
\subsection{Settings}\label{sec51}
This experiments was conducted on top-1 and top-5 predictions in 100 classes using the CIFAR-100 dataset \cite{cifar100}, which contains 60,000 RGB images, with 50,000 samples for training and 10,000 for testing. In this experiment, 5,000 training samples were used as the validation set to monitor top-1 or top-5 accuracies across all epochs for early-stopping.

We used a ResNet18 model \cite{resnet} and DenseNet121 model \cite{densenet} to train.
During training, we applied the following data augmentation techniques, random horizontal flips, random crops of size 32 $\times$ 32 on the images zero-padded with 4 pixels on each side, and either Cutout \cite{cutout} (with size 8 $\times$ 8) or Mixup \cite{mixup}. We selected the same learning schedule criteria and hyperparameters of optimization with \cite{cutout}; we used SGD with Nesterov momentum of 0.9, and weight decay of $5\times 10^{-4}$, and  trained for 200 epochs with 128 images per mini batch. The learning rate was initially set to 0.1 and scheduled to decrease by a factor of 5x after each 60th, 120th, and 160th epoch.

We evaluated the test accuracies of models both for early-stopping by top-1 validation accuracy and by top-5 validation accuracy. Note that we used two criteria for validation corresponding to the two target metrics of top-1 accuracy and top-5 accuracy respectively. We compared top-1 and top-5 accuracies and other top-k accuracies with different $k$s for better understanding of each loss. Each result is the average over ten trials with different random seeds.

\subsection{Results}\label{sec52}

\begin{table*}[!t]
\renewcommand{\arraystretch}{1.3}
\caption{Test accuracies (\%) on CIFAR-100 dataset. All reported accuracies are averaged over 10 different trials, and values in braces are standard deviations given as uncertainties. Results written in boldface are best ones for each performance metric.}
\label{tab:cifar100_result}
\centering
\begin{tabular}{c|ccccccccccc}
\hline
&\multicolumn{5}{c}{ResNet18}&&\multicolumn{5}{c}{DenseNet121}\\
&\multicolumn{2}{c}{with Cutout}&&\multicolumn{2}{c}{with Mixup}
        &&\multicolumn{2}{c}{with Cutout}&&\multicolumn{2}{c}{with Mixup}\\
Method & Top-1 & Top-5 && Top-1 & Top-5 && Top-1 & Top-5 && Top-1 & Top-5\\
\hline \hline
Cross entropy & \textbf{75.88(0.23)} & 93.11(0.16) && \textbf{76.76(0.23)} & 92.89(0.13)
                    && \textbf{79.05(0.36)} &\textbf{ 94.82(0.15)} && \textbf{79.32(0.25)} & 94.71(0.15) \\
Smooth top-5 SVM \cite{smoothTopK} & 67.31(0.31) & \textbf{93.76(0.12)} && N/A & N/A
                    && 73.55(0.36)& 94.50(0.19) && N/A & N/A \\
Top-5 grouping & 59.99(0.47)&92.64(0.13) && 57.66(0.52) & 92.52(0.21)
                    && 58.69(0.19) & 93.77(0.19) && 56.68(0.94) & 93.58(0.20) \\
Top-5 transition & 69.06(0.30)&93.62(0.18) && 74.26(0.23)&\textbf{94.07(0.14)}
                    && 71.92(0.26) & 94.38(0.24) && 76.28(0.28) & \textbf{95.18(0.06)} \\
\hline
\end{tabular}
\end{table*}

\begin{figure}[!t]
    \centering
    \includegraphics[trim=60 252 52 300,clip,width=3.5in]{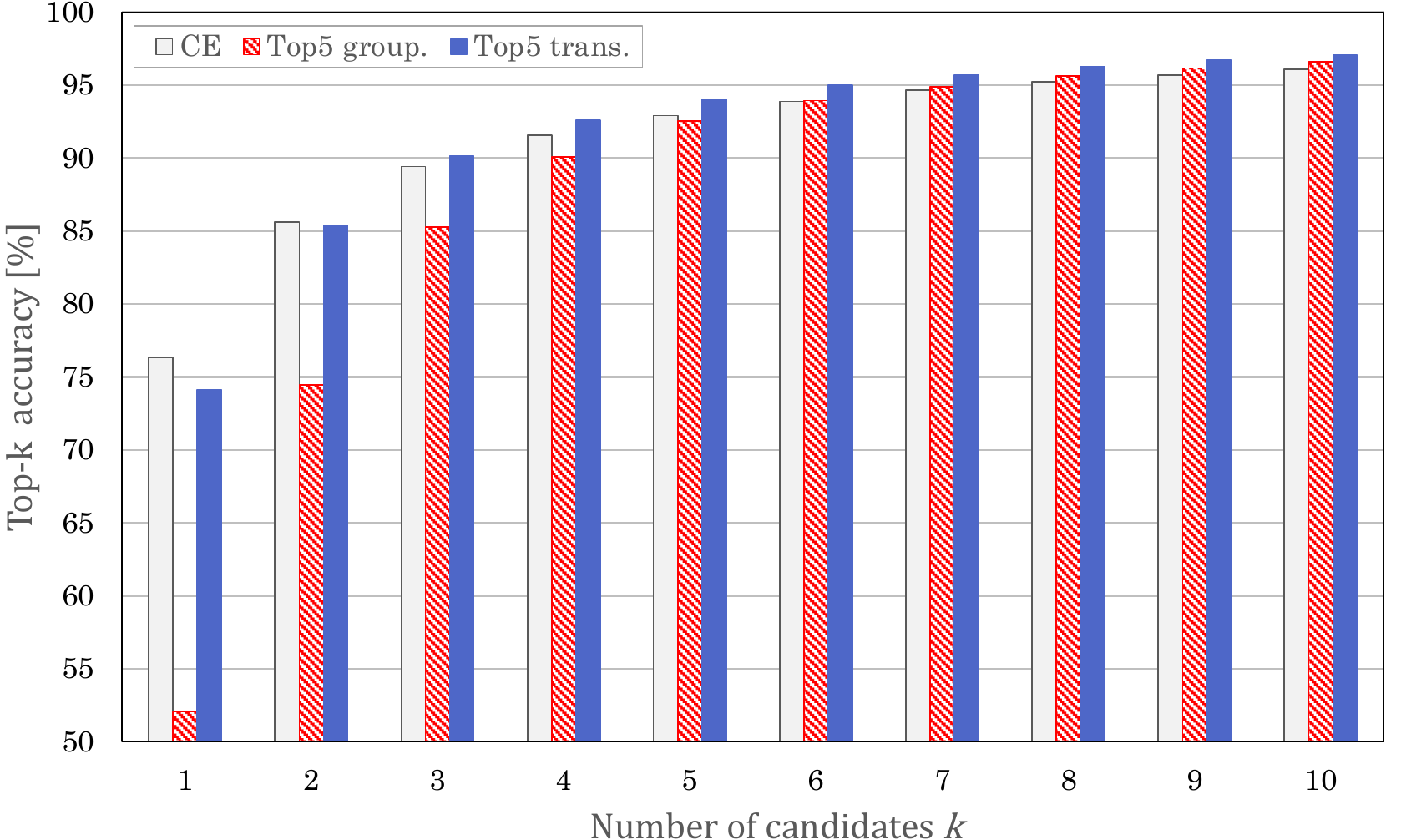}
    \caption{Test accuracy of top-k prediction for different $k$. The plot shows the average values over 10 trials of experiment using ResNet18 with Mixup augmentation.}
\label{fig:cifar100_hist}
\end{figure}

The main results are summarized in Table\,\ref{tab:cifar100_result}.
The table shows CE performed the best in top-1 accuracy, as we expected, while its top-5 accuracy was lower than our top-5 transition loss and smooth top-5 SVM loss \cite{smoothTopK} except for the result of DenseNet121 with Cutout. 
This indicates that trade-offs between top-1 and top-k accuracies occur even on an almost clean dataset, without adding label noises like Berrard \textit{et al.} \cite{smoothTopK}. 
Thus, CE is not always the best choice for training classifiers when we use not only their top-1 but also top-k predictions.

The table also shows that the best top-5 accuracy for each model was obtained by our top-5 transition loss with Mixup, where the improvement from CE was 0.9\,\% for ResNet18 and 0.36\,\% for DenseNet121. 
In the top-5 accuracy with Cutout, smooth top-5 SVM loss \cite{smoothTopK} has shown the slightly higher performance than our top-5 transition loss.
Smooth top-5 SVM loss cannot be directly applied to training on soft label like Mixup, while the top-5 transition loss has better affinity to training techniques. 

The top-5 grouping loss performed worse than CE in both top-1 and top-5 accuracies. 
As we discussed in the synthetic experiments, we consider the reason is that, in CIFAR-100 dataset, inter-class overlap are so small that the top-5 grouping loss cannot select robust decision boundaries.
As a result, the top-5 grouping loss performs worse than CE even in top-5 accuracy. 
In contrast, the top-5 transition loss has realized better top-5 performance than CE and the top-5 grouping loss. 
Difference between the transition loss and the top-5 grouping loss is to make transition from CE during optimization.
Therefore, we consider the transition successfully adds a bias to find robust decision boundaries to the top-5 grouping loss. 

Fig.\,\ref{fig:cifar100_hist} shows top-1 to top-10 accuracies from the results with ResNet18 and Mixup. 
CE loss gives the best top-k accuracy for $k \leq 2$ but defeated by the top-5 transition loss for $k \geq 3$ and also by top-5 grouping for $k \geq 6$.
These turning points consistently appear in the other conditions. 
In the case with DenseNet121 and Cutout CE was the best still in $k=5$, but defeated by the top-5 transition loss for $k \geq 9$ and by the top-5 grouping loss for $k \geq 11$. 
We found the top-5 transition loss gave consistently better performance for $k > 10$ than other losses in all cases.
When focusing on fixed target accuracy, the top-5 transition loss can realize 99\,\% with number of prediction candidates $k=25$ saving 8 compared with CE.
This means that using the transition loss reduces the number of candidates for achieving target accuracy, which is preferable characteristic in some practical applications. 

Mechanism behind the trade-offs on CIFAR-100 dataset can not be explained only by the relationship between model complexity and data distribution stated in Sec. \ref{sec44}. 
This is because that top-1 accuracy on the training set reached almost one and thus the models seem to have sufficient complexities to fit top-1 prediction of data distribution.
As discussed in previous paragraphs, the trade-offs also occur in deep learning regime. 
We observed them as the difference in generalization error, not in fitting error caused by insufficiency of model complexity. 
Further study is needed to understand what makes trade-offs occur in use of deep neural networks.

\section{Conclusion}\label{sec6}
In this work, we have presented an investigation for the cases that CE fails to obtain the best top-k accuracy of possible predictions.
Furthermore we have modified CE to improve top-k accuracy under such conditions. A key concept of the modification is to alternate top-k components of original softmax outputs with their summation as a single class. Based on this concept, we have proposed a top-k transition loss with an adaptive transition from normal CE in order to obtain a robust decision boundary.

Our experiments demonstrated that CE is not always the best choice to learn top-k prediction. 
In a series of experiments using synthetic datasets, we found that CE does not perform well in top-k accuracy when we have complex data distribution for a model to fit.
Specifically, it happens when distribution of a certain class has multiple modes that leads to complex decision boundary to achieve for optimal top-1 prediction.

We have compared top-k accuracies for each loss on the public CIFAR-100 dataset with ResNet18 and DenseNet121 as deep learning examples.
This experiments were conducted on top-1 and top-5 predictions in 100 classes.
While CE performed the best in top-1 accuracy, in top-5 accuracy our loss and the existing top-k loss performed better than CE except using DenseNet121 model and Cutout augmentation. 
The transition loss has been found to provide consistently better top-k accuracies than CE for $k > 10$. 
As a result, a ResNet18 model trained with our loss reached 99\,\% accuracy with $k=25$ candidates, which is a smaller candidate number than that of CE by 8.








\bibliographystyle{IEEEtran}

\bibliography{relatedworks}

%
%

\end{document}